\documentclass[11pt,a4paper]{article}
\usepackage[hyperref]{acl2021}
\usepackage{times}
\usepackage{latexsym}
\usepackage{subfigure}
\usepackage{graphicx}
\usepackage{booktabs}   
\usepackage{bm}

\usepackage{xcolor}

\def\aclpaperid{3291} 

\usepackage{microtype}

\aclfinalcopy 
\def\aclpaperid{3291} 

\title{Improving Speech Translation by Understanding and Learning from the Auxiliary Text Translation Task}

\author{Yun Tang, Juan Pino, Xian Li, Changhan Wang, Dmitriy Genzel \\
 Facebook AI\\
\texttt{\{yuntang,juancarabina,xianl,changhan,dgenzel\}@fb.com}         
}

\date{}

\begin{document}
\maketitle
\begin{abstract}
Pretraining and multitask learning are widely used to improve the speech to text translation performance.
In this study, we are interested in training a speech to text translation model along with an auxiliary text to text translation task. We conduct a detailed analysis to understand the impact of the auxiliary task on the primary task within the multitask learning framework. 
Our analysis confirms that multitask learning tends to generate similar decoder representations from different modalities and preserve more information from the pretrained text translation modules. We observe minimal negative transfer effect between the two tasks and sharing more parameters is helpful to transfer knowledge from the text task to the speech task.
The analysis also reveals that the modality representation difference at the top decoder layers is still not negligible, and those layers are critical for the translation quality.  
Inspired by these findings, we propose three methods to improve translation quality.
First, a parameter sharing and initialization strategy is proposed to  enhance information sharing between the tasks. Second, a novel attention-based regularization is proposed for the encoders and pulls the representations from different modalities closer. Third, an online knowledge distillation is proposed to enhance the knowledge transfer from the text to the speech task.
Our experiments show that the proposed approach improves translation performance by more than 2 BLEU over a strong baseline and achieves state-of-the-art results on the \textsc{MuST-C} 
English-German, English-French and English-Spanish language pairs. 
\end{abstract}

\section{Introduction}
End-to-end methods have achieved significant progress in speech to text translation (ST) and  even surpassed the traditional pipeline-based methods in some applications~\cite{Niehues2019TheI2,salesky2020phone}.
However, the success of end-to-end methods relies on large amounts of training data, which is quite expensive to obtain and relatively small in practice. Building ST systems from pretrained models with multitask learning (MTL) is widely used to overcome the limited training data issue~\cite{Weiss2017SequencetoSequenceMC,Anastasopoulos2018TiedML,Bahar2019ACS,Indurthi2020EndendST,Wang2019BridgingTG,Li2020MultilingualST}. 
Nevertheless, little prior work has been devoted to understanding the interactions between different tasks. \citet{Standley2019WhichTS} conduct an empirical study on computer vision tasks for MTL. 
They find many ``assumptions" for MTL may not be held for specific applications. For example, ``similar" tasks do not necessarily train better together. 

In this study, we focus on training the ST model along with an auxiliary text to text machine translation (MT) task. 
We are interested in the task interactions with different modalities and in improving the primary ST task with the help from the auxiliary MT task.
The model is initialized with pretrained modules from automatic speech recognition~(ASR) and MT. 
Two types of analysis are conducted on the fine-tuned 
multitask learned models. 
The first focuses on the model variation 
by comparing fine-tuned models with pretrained models for different tasks.  The second aims to measure internal representation differences due to different modalities.
The analysis leads to three main findings. First, the analysis confirms that MTL tends to generate similar model representations for different input modalities and preserves more information from the pretrained MT modules. Second, we do not observe significant negative transfer effect from the MT task to the corresponding ST task. Sharing more parameters is helpful  to transfer knowledge to the primary ST task. 
Finally, the top layers in the ST decoder are more critical to the translation performance and they are also more sensitive to the modality difference. The model representations from different modalities demonstrate larger difference for the top layers in our analysis.  

Inspired by these findings, we propose three techniques to enhance the performance of the primary ST task. 
First, we propose to maximize  parameter sharing 
between the ST and MT tasks, i.e.\ the entire decoder and the top encoder layers. Those shared parameters are initialized with the corresponding MT models. Second, a cross-attentive regularization is introduced for the encoders. It minimizes the $L2$ distance between two reconstructed encoder output sequences and
encourages the encoder outputs from different modalities to be closer to each other. Finally, an online knowledge distillation learning is introduced for MTL in order to enhance knowledge transfer from the MT to the ST task.  

Our contributions are summarized as follows:
\begin{enumerate}
    \item A detailed analysis is conducted on the interaction between the primary ST task and the auxiliary MT task.
    \item A parameter sharing and initialization strategy are proposed to encourage information sharing between tasks.
    \item Cross-attentive regularization and online knowledge distillation are proposed to reduce the model representation difference between different modalities and enhance the knowledge transfer from the MT task to the ST task.
    \item Our system achieves state of the art results on the \textsc{MuST-C} English-German (EN-DE), English-French (EN-FR) and English-Spanish (EN-ES) language pairs, with 2 or more BLEU gains over strong baselines.
\end{enumerate}



\section{Related Work}\label{sec:rel_work}
\textbf{Multitask learning} aims to improve generalization by leveraging domain-specific information contained in the training signals of related tasks~\cite{Vandenhende2020MultiTaskLF}. Compared with single task, MTL has many advantages, such as 
the potential to improve performance by sharing complementary information or acting as a regularizer. Many previous works focus on learning a good model for all tasks. \citet{Chen2018GradNormGN} study the gradients from different tasks and conduct task dependent gradient normalization to encourage different tasks to learn at similar speed.
\citet{Maninis2019AttentiveSO,Liu2019EndToEndML,Pfeiffer2020MADXAA} introduce task-dependent components to enhance individual task performance.

\citet{Weiss2017SequencetoSequenceMC} explore different multitask training strategies for ST, and they find the one-to-many strategy, in which an encoder is shared between the ST and ASR tasks,  is more effective. \citet{Anastasopoulos2018TiedML} further extend it to a triangle structure by concatenating ASR and ST models. \citet{Bahar2019ACS} compare different multitask strategies for the ST task, and they confirm many-to-one strategy, in which MT and ST are trained together and the decoder is shared between two tasks, is effective if extra bitext data is used. In this work, we carefully study the relation between co-trained tasks in the many-to-one strategy, and the analysis results guide us to propose three techniques to learn more from the auxiliary MT task and enhance the ST performance further.



\begin{figure}
\centering
\includegraphics[width=1.00\columnwidth]{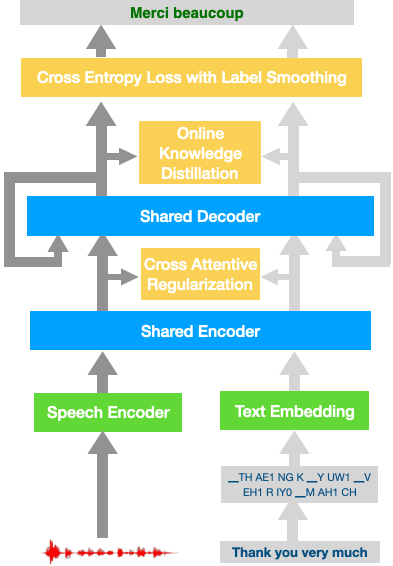}
\caption{Joint Training framework. The speech to text translation task is depicted as dark gray line, text to text translation task is illustrated as light gray line. The parameters in blue modules are shared between two tasks.}
\label{fig:jt_framework}
\end{figure}

\noindent\textbf{Model analysis}~\citet{Chatterji2020TheIR} propose criticality analysis to measure the importance of different modules from the trained model. 
Parameters in the selected module or layer are partially rolled back to the initial values, and the module criticality or importance is measured by the performance drop after modification. Larger performance drops indicate a more critical module. Inspired by their work, we extend it to the analysis on the jointly trained models with different pretrained modules and schemes.
\citet{Raghu2017SVCCASV,Morcos2018InsightsOR} propose to employ canonical correlation to measure the similarity between different models given the same input. 
We extend their work to study a model with inputs from different modalities. 

\section{Methods}\label{sec:methods}
The proposed ST system is co-trained with the MT task as depicted in \autoref{fig:jt_framework}. The modules in the primary ST task are connected with dark gray lines and the auxiliary MT task is illustrated with light gray lines.  The parameters in the blue modules are shared between the two tasks. During inference with speech input, only  modules related to the ST task are used.

The model has two encoders, a text encoder and a speech encoder, to take text  and speech input respectively. 
The decoder is shared between the two tasks. 
To encourage knowledge sharing between the two tasks, the top encoder layers are also shared. The parameters of the shared modules 
are initialized with a pretrained MT model. 
A novel cross-attentive regularization is proposed to reduce the distance between encoder outputs from different input modalities. 
We also introduce a novel online knowledge distillation method where the output from the auxiliary MT task is used to guide the ST model training. The cross-attentive regularization and online knowledge distillation are illustrated as orange modules in \autoref{fig:jt_framework} and the details are presented in the following two subsections.

\subsection{Cross-Attentive Regularization}\label{sec:car}
The cross-attentive regularization (CAR) is proposed to increase the similarity between the text encoder outputs and their corresponding speech encoder outputs.
Hence, the performance of the more difficult ST task can be improved by learning from the relatively easier MT task.
Encoder output sequences from different modalities can not be compared directly since they have different lengths. 
In CAR, the two reconstructed sequences are calculated from the text output sequence via self-attention or the speech output sequence via cross attention over the text output sequence. The two reconstructed sequences have the same length and the distance is simply measured as the $L2$ distance between the two sequences. 

Formally, we denote a speech to text translation 
training  sample as a triplet 
$o=(\mathbf{X^s}, \mathbf{x^t}, \mathbf{y})$. $\mathbf{X^s}\in\mathcal{R}^{d_s \times N}$, $\mathbf{x^t}\in\mathcal{R}^{M}$, and $\mathbf{y}\in\mathcal{R}^{K}$ are the speech feature input, text token input and target text output respectively. $N$, $M$ and $K$ are the corresponding sequence lengths.  
Assume  $\mathbf{H^s}=(\mathbf{h}^s_1,\mathbf{h}^s_2,\cdot\cdot\cdot,\mathbf{h}^s_N)$ and  $\mathbf{H^t}=(\mathbf{h}^t_1,\mathbf{h}^t_2,\cdot\cdot\cdot,\mathbf{h}^t_M)$, $ \mathbf{h}^s_n, \mathbf{h}^t_m \in \mathcal{R}^{d_h}$ are outputs from the speech encoder and text encoder respectively,  where $d_h$ is the dimension of the output  states. 
A similarity matrix $\mathbf{S} \in \mathcal{R}^{N\times M}$ is defined as the cosine distance between the tensors in the two sequences: 
\begin{equation}
s_{i,j} = \frac{(\mathbf{h}^s_i)' \cdot \mathbf{h}^t_j}{ {|| \mathbf{h}^s_i ||}_2 {|| \mathbf{h}^t_j ||}_2}
\end{equation}\label{equ:sim}
where $s_{i,j}$ is the $i$th row and $j$th column component in $\mathbf{S}$. 
The text encoder outputs $\mathbf{H}^t$ are reconstructed through the speech encoder outputs $\mathbf{H}^s$ and similarity matrix $\mathbf{S}$ as below.
\begin{equation}
    \mathbf{H}^{s\rightarrow t} = \mathbf{H}^s\cdot \mathrm{softmax}(\mathbf{S}) 
\end{equation}\label{equ:t2s}
$\mathbf{H}^{t\to t}$, the reconstruction  of $\mathbf{H}^t$ from itself, can be computed similarly via self-attention. 
CAR is defined as the $L2$ distance between the two reconstruction encoder outputs:
\begin{equation}
    \mathcal{L}_{CAR}(\theta_s)=\frac{1}{M} \left\Vert \mathbf{H}^{s\rightarrow t} - sg[\mathbf{H}^{t\rightarrow t}] \right\Vert_2 
\end{equation}
where $sg[\cdot]$ is the  stop-gradient  operator and $\theta_s$ are the ST model parameters. By optimizing the model with CAR, the speech encoder is encouraged to learn from more accurate text encoder and generates similar encoder outputs after reconstruction.
CAR is inspired by the attention mechanism between the encoder and decoder where the decoder states are reconstructed through encoder output states via the attention mechanism. 

\subsection{Online Knowledge Distillation}\label{sec:kd}
Knowledge distillation (KD) is widely used for model compression~\cite{Hinton2015DistillingTK,Kim2016SequenceLevelKD} where a smaller student network is trained to mimic the original teacher network by minimizing the loss between the student and teacher outputs. 
The ST task is considerably more difficult than the MT task since the speech input is noisier and more ambiguous than the text input. The accuracy of the MT model is usually much higher than the corresponding ST model. Knowledge distillation from a well trained MT model to a ST model has been proved to be an effective way to improve the ST performance~\cite{Liu2019EndtoEndST,gaido2020endtoend}.  
In this work, we extend knowledge distillation to the MTL framework where both ST and MT are fine-tuned simultaneously with shared parameters. 

Concretely, we assume an MTL model learns from a data set $\mathcal{D}$ with target vocabulary size $|V|$. The training criterion is to minimize negative log likelihood (NLL) for each example $o=(\mathbf{X^s}, \mathbf{x^t}, \mathbf{y}) \in \mathcal{D}$ from the training data:
\begin{eqnarray}
    \mathcal{L}_{NLL}(\theta_s)  &=&  -\sum_{o}^{\mathcal{D}} \sum_{k=1}^K \sum_{v=1}^{|V|} \delta(y_k=v) \nonumber \\
     && \log p(y_k=v|y_{<k},\mathbf{X}^s,\theta_s) 
\end{eqnarray}
where $\delta(\cdot)$ is the indicator function and $p$ the distribution from the ST model (parameterized by $\theta_s$).

Assume the probability distribution for  $y_k$ given text input  $\mathbf{x}^t$ and MT model $\theta_t$ is $q(y_k=v|y_{<k},\mathbf{x}^t,\theta_t)$, the knowledge distillation loss is defined as minimizing the cross-entropy with the MT’s probability distribution
\begin{eqnarray}
    \mathcal{L}_{KD}(\theta_s)  &=&  -\sum_{o}^{\mathcal{D}} \sum_{k=1}^K \sum_{v=1}^{|V|} q(y_k=v|y_{<k},\mathbf{x}^t,\theta_t) \nonumber \\
      && \log p(y_k=v|y_{<k},\mathbf{X}^s,\theta_s) 
\end{eqnarray}

The overall loss is the combination of  cross-attentive regularization,  knowledge distillation loss, negative log likelihood loss for both ST and MT, as follows:
\begin{eqnarray}\label{equ:total_cost}
    \mathcal{L}(\theta_s, \theta_t) &=& \alpha\mathcal{L}_{NLL}(\theta_s) + (1-\alpha)\mathcal{L}_{KD}(\theta_s) \nonumber\\
    && + \lambda\mathcal{L}_{CAR}(\theta_s)+ \mathcal{L}_{NLL}(\theta_t) 
\end{eqnarray}
where $\alpha$ and $\lambda$ are predefined hyper-parameters.

\section{Experimental Setup}\label{sec:setup}
Experiments are conducted on three \textsc{MuST-C}~\cite{Gangi2019MuSTCAM} language pairs: EN-DE, EN-ES and EN-FR. The models are developed and analyzed on the dev set and the final results are reported on the tst-COMMON set. We use WMT parallel data from different years, 2013 for Spanish, 2014 for German, and 2016 for French, 
as extra text training corpus for MTL. Case-sensitive detokenized BLEU is reported by \textsc{sacrebleu} with default options~\cite{post-2018-call}.

We use the ``T-Md" configuration 
from~\cite{Wang2020fairseqSF} in all experiments. The speech encoder has 12 transformer layers while the  decoder is with 6 transformer layers. For the MTL model, the text encoder has 6 transformer layers. 
The transformer layer has an input embedding size of 512 and middle layer dimension 2048.  We share parameters of all 6 text encoder transformer layers with the top 6 transformer layers in the speech encoder, hence both encoders use the same modules to generate the encoder outputs.

The Adam optimizer~\cite{kingma2014adam} with a learning rate 0.002 is employed in the experiments. Label smoothing and dropout rate are both set to 0.1. We choose $\alpha = 0.8$ and $\lambda = 0.02$  in \autoref{equ:total_cost} through grid search ($[0.1, 1.0]$ for $\alpha$ and $[0.01,0.05]$ for $\lambda$).

Input speech is represented as 80D log mel-filterbank coefficients computed every 10ms with a 25ms window.  Global channel mean and variance normalization is applied. The SpecAugment~\cite{park2019specaugment} data augmentation with the LB policy is applied in all experiments.
The input text tokens are converted into their corresponding pronunciation form as phoneme sequences~\cite{Tang2020AGM,Renduchintala2018MultiModalDA}. The grapheme to phoneme conversion is done through the ``g2p\_en'' python package~\cite{Lee2018LearningPF}. 
The leading phoneme in a word is appended with an extra ``\_" to mark word boundaries. In total, the vocabulary size for the input phonemes is 134. The target vocabulary consists of 10k ``unigram'' subword units learned by SentencePiece~\cite{Kudo2018SentencePieceAS} with full character coverage of all training text data. 

All ST or jointly trained models  are initialized with pretrained ASR  and MT modules. The ASR model is trained on the same English speech training data from \textsc{MuST-C} with the ``T-Md" configuration too.
The pretrained MT models are trained for each language pair with the aforementioned WMT data. The MT encoder and decoder configurations are the same as the text encoder and decoder in the MTL model mentioned above.

The models are fine-tuned to 100 epochs using 8 V100 GPUs for approximate one day. The batch size is 10,000 frames for speech to text translation samples and 10,000 tokens for parallel text samples per GPU. The model parameters are updated every $4$ batches. Speech training samples and text input samples are used to update the model alternatively. 
The models are trained with \textsc{Fairseq}~\cite{Ott2019fairseqAF,Wang2020fairseqSF}. 
The last 10 checkpoints are averaged for inference with beam size 5. \footnote{The source code will be released at https://github.com/pytorch/fairseq/tree/master/examples/speech\\\_text\_joint\_to\_text}. 

\begin{table}[t]
\centering
\begin{tabular}{ c | c | c | c  } 
 \hline
 Model  &	 \multicolumn{3}{|c}{Encoder}  \\ 
 \cline{2-4}
 Configuration & Speech & Text & Shared  \\
 \hline\hline
 ST	 &	ASR & None & None \\ 
 JT	 &	ASR & MT & None \\
 JT-S-ASR	 &	ASR & MT & ASR \\
 JT-S-MT	 &	ASR & MT & MT  \\
 \hline
\end{tabular}
\caption{Model initialization schemes}
\label{table:mod_ini}
\end{table}

\section{MTL Analysis}\label{sec:analysis}
 

\subsection{Model Variation} 
We extend \citet{Chatterji2020TheIR}'s work to analyze a MTL model.  We initialize models with different pretrained modules and fine-tune them for ST and MT tasks within the MTL framework.  The pretrained modules come from ASR and MT tasks.  

Criticality analysis is conducted on the ST model after the MTL fine-tuning step. The parameters in the selected modules are interpolated with corresponding parameters in the pretrained modules. \textsc{MuST-C} EN-DE dev set is used for BLEU computation.   
With different interpolation ratios, we obtain different BLEU scores.
The BLEU difference comes from two sources. The first one comes from the selected module itself. If the module is important and sensitive, very small perturbation could result in a nontrivial BLEU difference as~\cite{Chatterji2020TheIR}. Another source of difference is that if the selected module changes significantly to adapt to the ST task, rewinding the parameters back to the initial task may lead to a substantial decrease in BLEU. We attempt to quantify the extent of the degradation from the second source, which can be indicative of the model variation from the pretrained task to the ST task. This is accomplished by comparing the BLEU differences for the same module but using different initialization and training schemes.  

\autoref{table:mod_ini} lists models initialized with different pretrained modules. 
``ST" designates a ST model trained with the single ST task, ``JT" corresponds to a ST model trained with the primary ST task and auxiliary MT task together.  
``JT-S-ASR" and ``JT-S-MT" are another two jointly trained models but with the top encoder layers shared as described in \autoref{sec:setup}. 
The difference between the two models is how we initialized the shared encoder layers, either from the pretrained ASR model for ``JT-S-ASR" or from the pretrained MT model for ``JT-S-MT".

\begin{figure}
\hfill
\subfigure[ST Enc.]{\includegraphics[width=.48\columnwidth]{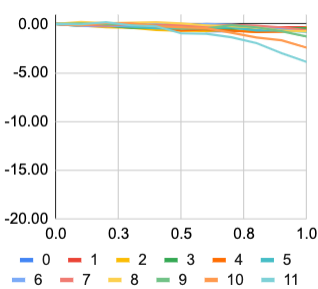}}
\hfill
\subfigure[ST Dec.]{\includegraphics[width=.48\columnwidth]{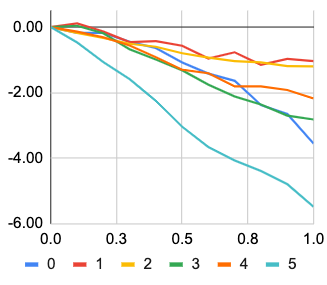}}
\hfill
\caption{Criticality analysis for the ``ST" model.}
\label{fig:imp_st}
\end{figure}

\noindent\textbf{ST}~ \autoref{fig:imp_st} shows the analysis for the ``ST" model. The x-axis is the interpolation ratio and ``1.0" means the pretrained parameters are used. The y-axis is the relative change in BLEU compared with the well-trained ST model. \textbf{It is clear that higher layers are more critical to the performance}. Around 5 BLEU decrease is observed on the top encoder layer~(11) and top decoder layer~(5) during the criticality tests. The following analysis will compare with \autoref{fig:imp_st} and we can separate the aforementioned second source from the first one.   

\begin{figure}
\hfill
\subfigure[JT Enc.]{\includegraphics[width=.48\columnwidth]{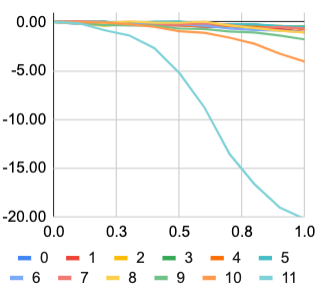}}
\hfill
\subfigure[JT Dec.]{\includegraphics[width=.48\columnwidth]{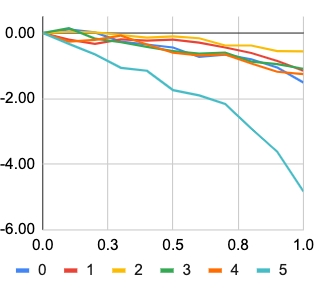}}
\hfill
\caption{Criticality analysis for the ``JT" model.}
\label{fig:imp_jt}
\end{figure}

\noindent\textbf{JT}~ \autoref{fig:imp_jt} presents the analysis for the ``JT" model. The jointly trained model shows smaller degradation compared with ``ST" for the decoder layers. \textbf{This indicates that training the ST and MT tasks together helps to preserve more information from the original MT decoder and partially remedies the catastrophic forgetting~\cite{McCloskey1989CatastrophicII} during the fine-tuning phase.}
On the other hand, after rolling parameters back to the initial ASR model, the jointly trained model shows a larger degradation for the encoder layers. This means that the speech encoder in the jointly trained model has deviated far away from the speech encoder in the initial ASR task. We conclude that the shared decoder is subject to more constraints since it is optimized toward both MT and ST tasks while the speech encoder has to undergo larger changes in order to align with the text encoder, although there is no parameter sharing between two encoders. 

\begin{figure}[t]
\hfill
\subfigure[JT-S-ASR Enc.]{\includegraphics[width=.48\columnwidth]{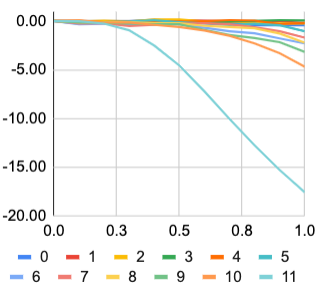}}
\hfill
\subfigure[JT-S-ASR Dec.]{\includegraphics[width=.48\columnwidth]{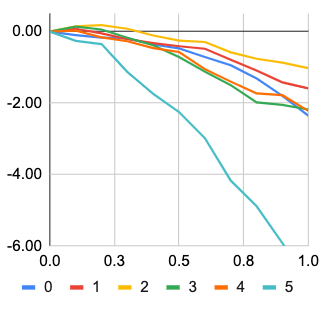}}
\hfill
\caption{Criticality analysis for the ``JT-S-ASR" model. The shared encoder layers are initialized with the layers from the ASR encoder.}
\label{fig:imp_jt_asr}
\end{figure}

\begin{figure}[t]
\hfill
\subfigure[JT-S-MT Enc.]{\includegraphics[width=.48\columnwidth]{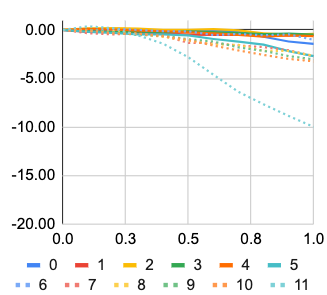}}
\hfill
\subfigure[JT-S-MT Dec.]{\includegraphics[width=.48\columnwidth]{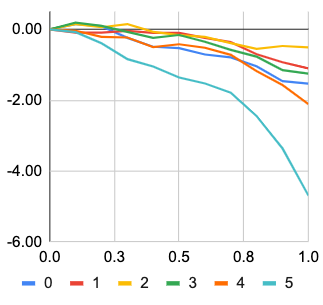}}
\hfill
\caption{Criticality analysis for the ``JT-S-MT" model. The shared encoder layers are initialized with the layers from the MT encoder.}
\label{fig:imp_jt_mt}
\end{figure}

\noindent\textbf{JT-S-ASR and JT-S-MT}~
Results for models with the top encoder layers shared are presented in \autoref{fig:imp_jt_asr} and \ref{fig:imp_jt_mt}.  In ``JT-S-MT", the top 6 shared encoder layers are initialized with the pretrained MT encoder. We illustrate their BLEU difference trajectories with dotted lines in \autoref{fig:imp_jt_mt} (a) so they can be easily distinguished from other layers initialized from the ASR encoder. 

The BLEU  difference for the top encoder layer is down from 20.2 to 17.6 when the parameters are replaced with the ones in the pretrained ASR encoder. It is further reduced to 10.0 if the shared layers are initialized with  MT encoder layers. 
The BLEU differences in the decoder layers are mixed. The performance of ``JT-S-ASR" degrades quickly in the criticality test for the top decoder layer, while ``JT-S-MT performs similarly in the test as ``JT" decoder.  
We argue that the top layers in the fine-tuned ST encoder might be closer to the MT encoder than the ASR encoder. 
\textbf{It preserves more information from the MT task by sharing more parameters between two tasks and initializing them with pretrained MT modules}. This is a desirable property since we want to transfer more knowledge from the text corpus to the ST task.


\subsection{Modality Variation}
The jointly trained model takes input from two modalities, i.e.\ text or speech, and we are interested in the model internal representation difference for paired inputs.
Given text target $\mathbf{y}$, we extract the decoder hidden state representations for the corresponding text input $\mathbf{x}_t$ and speech input $\mathbf{X}_s$.
The decoder representation difference solely comes from different input modalities. The difference is quantified by the correlation coefficient over all samples evaluated between two input modalities:
\begin{equation}
    r^{s,t}(l,d) = \frac{\sigma_{st}(l,d)}{\sigma_s(l,d)\sigma_t(l,d)} 
\end{equation}
where $\sigma_z(l,d), z\in [s,t]$ is the standard deviations of decoder hidden states at layer $l$ for component $d$ in all samples, and $\sigma_{st}(l,d)$ is the corresponding covariance.
The layer-wise correlation coefficient is the average of all components:
\begin{equation}
    r^{s,t}(l) = \frac{1}{D} \sum_d r^{s,t}(l,d) 
\end{equation}

\begin{figure}
\centering
\includegraphics[width=.7\columnwidth]{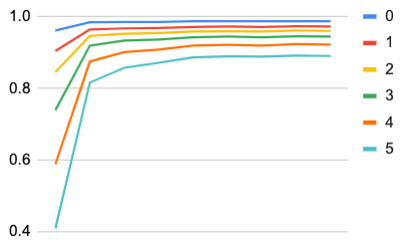}
\caption{Comparison of decoder layers correlation coefficients between text and speech input (``JT-S-MT").}
\label{fig:dec_coe_all}
\end{figure}

\autoref{fig:dec_coe_all} depicts the correlation coefficient between speech input and text input for each decoder layer in the model ``JT-S-MT". The x-axis is the number of training epochs and the y-axis represents the correlation coefficient for each layer. There are two observations. 
First, the correlation coefficients become larger and close to ``1.0" as training converges.   
Second, the higher the layer, the smaller the correlation coefficient. We hypothesize that the inputs to the lower layers are dominated by the decoder text embeddings, which are the same for both modalities, and the inputs to the higher layers would contain more information from the encoder outputs, which result in the decoder internal representation differences.
\textbf{The analysis shows a well trained MTL decoder  has similar representations for paired text and speech input. However, the top decoder layers still have nontrivial representation differences due to different modalities}.  


\section{Experimental Results}\label{sec:expt_rst}
\subsection{Main Results}
The main ST results are presented in \autoref{tab:ast}. The first three rows are results from the literature. ``ST" and ``JT" are models initialized as \autoref{table:mod_ini} and studied in \autoref{sec:analysis}.  
The last row (``JT Proposed") presents  results from the proposed system, in which the top encoder layers and decoder are shared, and the models are optimized following \autoref{equ:total_cost}. The second column (``\#pars(m)") lists the number of parameters used during inference.
From \autoref{tab:ast}, our ``ST" baseline is comparable to the previously reported results except~\cite{Pino2020SelfTrainingFE}, who use a much larger model and additional weakly supervised speech training data. As expected, the vanilla joint training baseline (``JT") outperforms the ``ST" baseline with the help of extra bitext training data. Finally, the proposed joint training model (``JT Proposed") achieves 2.0$\sim$2.7 BLEU gains over the strong joint training baseline (``JT").
\begin{table}
    \centering
    \small
    \begin{tabular}{l|c|c|c|c}
    \toprule
    {Data corpus} &\#pars(m)& {DE} & {ES} &{FR} \\
    \hline
    \citet{gangi2019onetomany} & 30& 17.7 &   20.9  &   26.5  \\
    \citet{Inaguma2020ESPnetSTAS} & - & 22.9 & 28.0 & 32.7 \\
    \citet{Pino2020SelfTrainingFE} & 435 & 25.2 & - & 34.5 \\
    \hline\hline
    ST & 76 & 21.5  &  28.1  & 33.8    \\
    JT & 76 & 24.1 &  29.0  &  35.1    \\
     \hline\hline
    JT Proposed & 76 & 26.8 & 31.0 & 37.4 \\
    \bottomrule
    \end{tabular}
    \caption{ BLEU on three language pairs in the MuST-C tst-COMMON datasets. }\label{tab:ast}
\end{table}

\subsection{Ablation}
\autoref{tab:ablation} breaks down the performance gains into individual components/changes.
Sharing encoder layers improves the quality for all three language pairs (``JT" v.s. ``JT-S-ASR"). Initializing the shared encoder layers with pretrained MT modules leads to BLEU increase for two of the three evaluated translation pairs (``JT-S-ASR" v.s. ``JT-S-MT"). For EN-FR, the degradation is minimal (-0.1 BLEU). Overall, sharing top encoder layers can increase BLEU  by 0.2$\sim$0.7 (``JT-S-MT" v.s. ``JT"). CAR further improves the translation by another 0.3$\sim$0.9 BLEU. The best results are achieved by applying the shared top encoder layers, CAR and online KD together. They are about 2.9+ BLEU better than the single task based system (``ST") and achieve 2+ BLEU increase on top of the strong vanilla joint training system(``JT").   

\begin{table}[!tbp]
    \centering
    \small
    \begin{tabular}{l|c|c|c}
    \toprule
      & EN-DE & EN-ES & EN-FR  \\
    \hline
    JT  & 24.1 &  29.0  &  35.1 \\
    \hline
    JT-S-ASR & 24.4 & 29.4 & 35.4 \\
    \hline
    \hline
    JT-S-MT & 24.7 & 29.7 & 35.3 \\
   $\;\;\;$+ CAR & 25.0 & 30.4  & 36.2 \\
   $\;\;\;$+ CAR + KD & 26.8 & 31.0 & 37.4\\
    \bottomrule
    \end{tabular}
    \caption{ Ablation study. }\label{tab:ablation}
\end{table}

\begin{figure} 
\hfill
\subfigure[JT Proposed Enc.]{\includegraphics[width=.48\columnwidth]{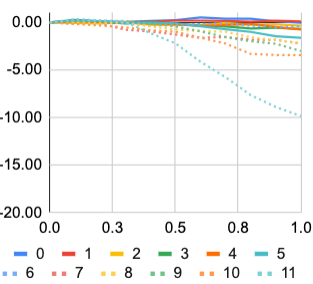}}
\hfill
\subfigure[JT Proposed Dec.]{\includegraphics[width=.48\columnwidth]{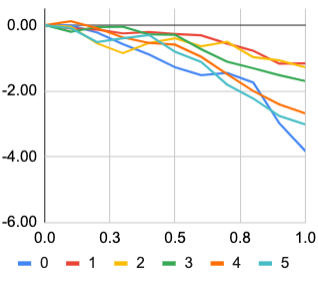}}
\hfill
\caption{Criticality analysis for ``JT Proposed". }
\label{fig:imp_jt_all}
\end{figure}

\begin{figure}
    \centering
    \includegraphics[width=1.05\columnwidth]{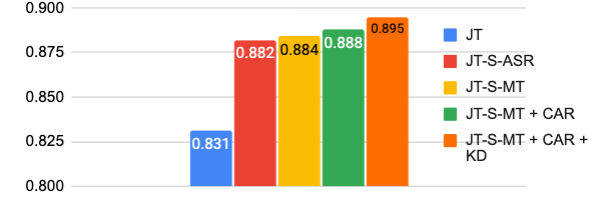}
    \caption{Correlation coefficient for the top decoder layers (epoch 100).}
    \label{fig:cor_coe}
\end{figure}
\autoref{fig:imp_jt_all} demonstrates the model variation for the proposed system on the \textsc{MuST-C} EN-DE dev set. Compared with \autoref{fig:imp_jt_mt}, 
the decoder shows less degradation during the criticality test and it shows CAR and online KD help to preserve more information from the MT task.
\autoref{fig:cor_coe} shows the corresponding correlation coefficients between paired text and speech input from the top decoder layer from different model configurations. It also confirms that the proposed methods, i.e., shared top encoder layers, CAR and online KD, all reduce the modality difference substantially. 

\subsection{Task Dependent Components}
In MLT, many works~\cite{Maninis2019AttentiveSO,Liu2019EndToEndML,Zhang2020ImprovingMM,Pfeiffer2020MADXAA} employ task-dependent components to alleviate the negative transfer effect. In \autoref{tab:task_dep_com}, we compare the ``JT-S-MT" model with two variants using different task-dependent components. 
The first one (``JT-S-MT + Adapter")~\cite{Bapna2019SimpleSA}  adds an extra adapter module on the top of the speech encoder. 
Hence, the speech encoder outputs, which are generated from shared encoder layers, are further processed to reduce the difference between speech input and text input.
The adapter module consists of a linear layer and layer normalization layer.  
The second variant (``JT-S-MT + Dedicated Attention")~\cite{Blackwood2018MultilingualNM} introduces dedicated decoder modules for different tasks. Attention layers between encoder and decoder,  and the layer normalization modules are not shared between the ST and MT tasks. It gives the decoder more flexibility to handle information from different modalities. 

The results show the extra adapter layer doesn't bring gain while the task dependent attention module actually makes the performance worse. It indicates that the negative transfer effect is not significant in this study and adding extra task-dependent components might not be necessary.  
\begin{table}[]
    \centering
    \small
    \begin{tabular}{c|c}
    \toprule
         Model & BLEU \\
         \hline
         JT-S-MT &  24.7 \\
         \hline
         JT-S-MT + Adapter & 24.7 \\
         \hline
         JT-S-MT + Dedicated Attention  & 24.2 \\
         \bottomrule
    \end{tabular}
    \caption{BLEU score for models with task dependent components}
    \label{tab:task_dep_com}
\end{table}

\subsection{Impact on the MT Task}~\label{sec:expt_mt}
As shown in~\autoref{tab:ast}, training ST models with an auxiliary MT task improves the translation quality substantially. It may be interesting to examine the impact on the auxiliary task itself. We evaluate the MT model jointly trained with the ST task.
Results are shown in \autoref{tab:mt}. ``ST (JT Proposed)" in the first row corresponds to the best results obtained for the ST task. The detailed experimental setup is described in \autoref{sec:app}.
For reference, we also include the MT evaluation results from \textsc{MuST-C}~\cite{Gangi2019MuSTCAM} in the second row. All MT models (in the last 4 rows) take phoneme sequences as input instead of SentencePiece. 

``MT" (row 3) shows the results from pretrained MT models on WMT. In the ``MT (Tuned)" row, the MT models pretrained on WMT are fine-tuned on the \textsc{MuST-C} datasets. The large improvements clearly show a domain mismatch between WMT and \textsc{MuST-C}. The MT models trained with WMT data are improved after fine-tuning, and they are comparable with the ones reported in ~\cite{Gangi2019MuSTCAM}, though the input token is in pronunciation form, which is more ambiguous than the corresponding SentencePiece unit. 

``MT (JT)" and ``MT (JT Proposed)" are results from the co-trained MT models in ``JT" and ``JT Proposed" respectively.
After fine-tuning using both MuST-C (speech and text) and WMT (text only) training data, the auxiliary MT models perform better than the corresponding ST models. 
The proposed techniques further improve the co-trained MT models by 0.7$\sim$1.6 BLEU. While this is a surprising result, we note that the dedicated MT models may be improved with better hyperparameter tuning. In conclusion, the results show the proposed methods are effective to unify two tasks into one model with minimal negative transfer effect.


\begin{table}[!tbp]
    \centering
    \small
    \begin{tabular}{l|c|c|c}
    \toprule
      & EN-DE & EN-ES & EN-FR  \\
    \hline
    ST (JT Proposed) & 26.8 & 31.0 & 37.4\\
   \hline
    MT~\cite{Gangi2019MuSTCAM} & 28.1 &  34.2  &  42.2 \\
   \hline
   MT & 25.4 & 27.7  & 33.5  \\ 
   MT (Tuned) & 29.6 & 34.3 & 41.4  \\
   \hline
   MT (JT) & 28.9 & 33.9 & 41.6 \\
   MT (JT  Proposed) & 30.5 & 34.7 & 42.3 \\
   
    \bottomrule
    \end{tabular}
    \caption{ Comparison between ST and MT. }\label{tab:mt}
\end{table}

\section{Conclusions}
In this study, we focus on understanding the interactions between the ST and MT tasks under the MTL framework, and on boosting the performance of the primary ST model with the auxiliary MT task. Two types of analysis on model variation and modality variation, are conducted on the MTL models. The analysis demonstrates MTL helps to preserve information from the MT task and generates similar model representations for different modalities. We observe a minimal negative transfer effect between the two tasks. Sharing more parameters can further boost the information transfer from the MT task to the ST model.
The analysis also reveals that the model representation difference due to modality difference is nontrivial, especially for the top decoder layers, which are critical for the translation performance.
Inspired by the findings, we propose three techniques to increase knowledge transfer from the MT task to the ST task. These techniques include parameter sharing and initialization strategy to improve the information sharing between tasks, CAR and online KD to encourage the ST system to learn more from the auxiliary MT task and then generate similar model representations from different modalities.
Our results show that the proposed methods improve translation performance and achieve state-of--the-art results on three \textsc{MuST-C} language pairs.
\bibliography{anthology,acl2021}

\begin{thebibliography}{38}
\expandafter\ifx\csname natexlab\endcsname\relax\def\natexlab#1{#1}\fi

\bibitem[{Anastasopoulos and Chiang(2018)}]{Anastasopoulos2018TiedML}
Antonios Anastasopoulos and David Chiang. 2018.
\newblock \href {https://arxiv.org/pdf/1802.06655.pdf} {Tied multitask learning
  for neural speech translation}.
\newblock In \emph{NAACL-HLT}.

\bibitem[{Bahar et~al.(2019)Bahar, Bieschke, and Ney}]{Bahar2019ACS}
Parnia Bahar, Tobias Bieschke, and Hermann Ney. 2019.
\newblock \href {http://arxiv.org/abs/1911.08870} {A comparative study on
  end-to-end speech to text translation}.
\newblock In \emph{ASRU}.

\bibitem[{Bapna et~al.(2019)Bapna, Arivazhagan, and Firat}]{Bapna2019SimpleSA}
Ankur Bapna, N.~Arivazhagan, and Orhan Firat. 2019.
\newblock \href {https://arxiv.org/pdf/1909.08478.pdf} {Simple, scalable
  adaptation for neural machine translation}.
\newblock In \emph{EMNLP/IJCNLP}.

\bibitem[{Blackwood et~al.(2018)Blackwood, Ballesteros, and
  Ward}]{Blackwood2018MultilingualNM}
G.~Blackwood, Miguel Ballesteros, and T.~Ward. 2018.
\newblock \href {http://arxiv.org/abs/1806.03280} {Multilingual neural machine
  translation with task-specific attention}.
\newblock In \emph{COLING}.

\bibitem[{Chatterji et~al.(2020)Chatterji, Neyshabur, and
  Sedghi}]{Chatterji2020TheIR}
Niladri~S. Chatterji, Behnam Neyshabur, and H.~Sedghi. 2020.
\newblock \href {http://arxiv.org/abs/1912.00528} {The intriguing role of
  module criticality in the generalization of deep networks}.
\newblock In \emph{ICLR}.

\bibitem[{Chen et~al.(2018)Chen, Badrinarayanan, Lee, and
  Rabinovich}]{Chen2018GradNormGN}
Z.~Chen, Vijay Badrinarayanan, Chen-Yu Lee, and Andrew Rabinovich. 2018.
\newblock \href {http://arxiv.org/abs/1711.02257} {Gradnorm: Gradient
  normalization for adaptive loss balancing in deep multitask networks}.
\newblock In \emph{ICML}.

\bibitem[{Gaido et~al.(2020)Gaido, Gangi, Negri, and
  Turchi}]{gaido2020endtoend}
Marco Gaido, Mattia Antonino~Di Gangi, Matteo Negri, and Marco Turchi. 2020.
\newblock \href {http://arxiv.org/abs/2006.02965} {End-to-end
  speech-translation with knowledge distillation: Fbk@iwslt2020}.

\bibitem[{Gangi et~al.(2019{\natexlab{a}})Gangi, Cattoni, Bentivogli, Negri,
  and Turchi}]{Gangi2019MuSTCAM}
Mattia Antonino~Di Gangi, Roldano Cattoni, Luisa Bentivogli, Matteo Negri, and
  Marco Turchi. 2019{\natexlab{a}}.
\newblock \href {https://www.aclweb.org/anthology/N19-1202.pdf} {{MuST-C}: a
  multilingual speech translation corpus}.
\newblock In \emph{NAACL-HLT}.

\bibitem[{Gangi et~al.(2019{\natexlab{b}})Gangi, Negri, and
  Turchi}]{gangi2019onetomany}
Mattia Antonino~Di Gangi, Matteo Negri, and Marco Turchi. 2019{\natexlab{b}}.
\newblock \href {http://arxiv.org/abs/1910.03320} {One-to-many multilingual
  end-to-end speech translation}.
\newblock In \emph{ASRU}.

\bibitem[{Hinton et~al.(2015)Hinton, Vinyals, and
  Dean}]{Hinton2015DistillingTK}
Geoffrey~E. Hinton, Oriol Vinyals, and J.~Dean. 2015.
\newblock Distilling the knowledge in a neural network.
\newblock \emph{ArXiv}, abs/1503.02531.

\bibitem[{Inaguma et~al.(2020)Inaguma, Kiyono, Duh, Karita, Soplin, Hayashi,
  and Watanabe}]{Inaguma2020ESPnetSTAS}
H.~Inaguma, S.~Kiyono, K.~Duh, S.~Karita, N.~Soplin, T.~Hayashi, and
  S.~Watanabe. 2020.
\newblock Espnet-st: All-in-one speech translation toolkit.
\newblock In \emph{ACL}.

\bibitem[{Indurthi et~al.(2020)Indurthi, Han, Lakumarapu, seok Lee, Chung, Kim,
  and Kim}]{Indurthi2020EndendST}
Sathish~Reddy Indurthi, HouJeung Han, Nikhil~Kumar Lakumarapu, Beom seok Lee,
  Insoo Chung, Sang-Ha Kim, and Chanwoo Kim. 2020.
\newblock End-end speech-to-text translation with modality agnostic
  meta-learning.
\newblock In \emph{ICASSP}.

\bibitem[{Kim and Rush(2016)}]{Kim2016SequenceLevelKD}
Yoon Kim and Alexander~M. Rush. 2016.
\newblock Sequence-level knowledge distillation.
\newblock In \emph{EMNLP}.

\bibitem[{Kingma and Ba(2014)}]{kingma2014adam}
Diederik~P Kingma and Jimmy Ba. 2014.
\newblock Adam: A method for stochastic optimization.
\newblock In \emph{ICLR}.

\bibitem[{Kudo and Richardson(2018)}]{Kudo2018SentencePieceAS}
T.~Kudo and J.~Richardson. 2018.
\newblock \href {http://arxiv.org/abs/1808.06226} {Sentencepiece: A simple and
  language independent subword tokenizer and detokenizer for neural text
  processing}.
\newblock In \emph{EMNLP}.

\bibitem[{Lee and Kim(2018)}]{Lee2018LearningPF}
Y.~Lee and T.~Kim. 2018.
\newblock Learning pronunciation from a foreign language in speech synthesis
  networks.
\newblock \emph{ArXiv}.

\bibitem[{Li et~al.(2020)Li, Wang, Tang, Tran, Tang, Pino, Baevski, Conneau,
  and Auli}]{Li2020MultilingualST}
Xian Li, Changhan Wang, Yun Tang, Chau Tran, Yuqing Tang, Juan Pino, Alexei
  Baevski, Alexis Conneau, and Michael Auli. 2020.
\newblock \href {http://arxiv.org/abs/2010.12829} {Multilingual speech
  translation with efficient finetuning of pretrained models}.
\newblock \emph{arXiv: Computation and Language}.

\bibitem[{Liu et~al.(2019{\natexlab{a}})Liu, Johns, and
  Davison}]{Liu2019EndToEndML}
Shikun Liu, Edward Johns, and A.~Davison. 2019{\natexlab{a}}.
\newblock \href {https://arxiv.org/pdf/1803.10704.pdf} {End-to-end multi-task
  learning with attention}.
\newblock \emph{2019 IEEE/CVF Conference on Computer Vision and Pattern
  Recognition (CVPR)}, pages 1871--1880.

\bibitem[{Liu et~al.(2019{\natexlab{b}})Liu, Xiong, He, Zhang, Wu, Wang, and
  Zong}]{Liu2019EndtoEndST}
Yuchen Liu, Hao Xiong, Zhongjun He, Jiajun Zhang, Hua Wu, Haifeng Wang, and
  Chengqing Zong. 2019{\natexlab{b}}.
\newblock \href {http://arxiv.org/abs/1904.08075} {End-to-end speech
  translation with knowledge distillation}.
\newblock In \emph{Interspeech}.

\bibitem[{Maninis et~al.(2019)Maninis, Radosavovic, and
  Kokkinos}]{Maninis2019AttentiveSO}
K.~Maninis, Ilija Radosavovic, and I.~Kokkinos. 2019.
\newblock \href {https://arxiv.org/pdf/1904.08918.pdf} {Attentive
  single-tasking of multiple tasks}.
\newblock \emph{2019 IEEE/CVF Conference on Computer Vision and Pattern
  Recognition (CVPR)}, pages 1851--1860.

\bibitem[{McCloskey and Cohen(1989)}]{McCloskey1989CatastrophicII}
M.~McCloskey and N.~J. Cohen. 1989.
\newblock Catastrophic interference in connectionist networks: The sequential
  learning problem.
\newblock \emph{Psychology of Learning and Motivation}, 24:109--165.

\bibitem[{Morcos et~al.(2018)Morcos, Raghu, and Bengio}]{Morcos2018InsightsOR}
Ari~S. Morcos, M.~Raghu, and S.~Bengio. 2018.
\newblock \href {https://arxiv.org/pdf/1806.05759.pdf} {Insights on
  representational similarity in neural networks with canonical correlation}.
\newblock In \emph{NeurIPS}.

\bibitem[{Niehues et~al.(2019)Niehues, Cattoni, St{\"u}ker, Negri, Turchi,
  Salesky, Sanabria, Barrault, Specia, and Federico}]{Niehues2019TheI2}
Jan Niehues, R.~Cattoni, Sebastian St{\"u}ker, Matteo Negri, Marco Turchi,
  Elizabeth Salesky, Ramon Sanabria, Lo{\"i}c Barrault, Lucia Specia, and
  Marcello Federico. 2019.
\newblock \href {https://doi.org/10.5281/zenodo.3532822} {The {IWSLT} 2019
  evaluation campaign}.

\bibitem[{Ott et~al.(2019)Ott, Edunov, Baevski, Fan, Gross, Ng, Grangier, and
  Auli}]{Ott2019fairseqAF}
Myle Ott, Sergey Edunov, Alexei Baevski, Angela Fan, S.~Gross, Nathan Ng, David
  Grangier, and M.~Auli. 2019.
\newblock \href {http://arxiv.org/abs/1904.01038} {fairseq: A fast, extensible
  toolkit for sequence modeling}.
\newblock In \emph{NAACL}.

\bibitem[{Park et~al.(2019)Park, Chan, Zhang, Chiu, Zoph, Cubuk, and
  Le}]{park2019specaugment}
D.~Park, W.~Chan, Y.~Zhang, C.~Chiu, B.~Zoph, E.~Cubuk, and Q.~Le. 2019.
\newblock \href {http://arxiv.org/abs/1904.08779} {Specaugment: A simple data
  augmentation method for automatic speech recognition}.
\newblock \emph{Interspeech}.

\bibitem[{Pfeiffer et~al.(2020)Pfeiffer, Vulic, Gurevych, and
  Ruder}]{Pfeiffer2020MADXAA}
Jonas Pfeiffer, Ivan Vulic, Iryna Gurevych, and Sebastian Ruder. 2020.
\newblock \href {https://arxiv.org/pdf/2005.00052.pdf} {{MAD-X}: An
  adapter-based framework for multi-task cross-lingual transfer}.
\newblock In \emph{EMNLP}.

\bibitem[{Pino et~al.(2020)Pino, Xu, Ma, Dousti, and
  Tang}]{Pino2020SelfTrainingFE}
J.~Pino, Q.~Xu, X.~Ma, M.~Dousti, and Y.~Tang. 2020.
\newblock \href {http://arxiv.org/abs/2006.02490} {Self-training for end-to-end
  speech translation}.
\newblock In \emph{Interspeech}.

\bibitem[{Post(2018)}]{post-2018-call}
Matt Post. 2018.
\newblock \href {https://doi.org/10.18653/v1/W18-6319} {A call for clarity in
  reporting {BLEU} scores}.
\newblock In \emph{Proceedings of the Third Conference on Machine Translation:
  Research Papers}, pages 186--191, Brussels, Belgium. Association for
  Computational Linguistics.

\bibitem[{Raghu et~al.(2017)Raghu, Gilmer, Yosinski, and
  Sohl-Dickstein}]{Raghu2017SVCCASV}
M.~Raghu, J.~Gilmer, J.~Yosinski, and Jascha Sohl-Dickstein. 2017.
\newblock \href {https://arxiv.org/abs/1706.05806} {Svcca: Singular vector
  canonical correlation analysis for deep learning dynamics and
  interpretability}.
\newblock In \emph{NIPS}.

\bibitem[{Renduchintala et~al.(2018)Renduchintala, Ding, Wiesner, and
  Watanabe}]{Renduchintala2018MultiModalDA}
A.~Renduchintala, S.~Ding, M.~Wiesner, and S.~Watanabe. 2018.
\newblock Multi-modal data augmentation for end-to-end asr.
\newblock In \emph{INTERSPEECH}.

\bibitem[{Salesky and Black(2020)}]{salesky2020phone}
Elizabeth Salesky and Alan~W Black. 2020.
\newblock \href {http://arxiv.org/abs/2005.13681} {Phone features improve
  speech translation}.
\newblock In \emph{ACL}.

\bibitem[{Standley et~al.(2020)Standley, Zamir, Chen, Guibas, Malik, and
  Savarese}]{Standley2019WhichTS}
T.~Standley, A.~Zamir, Dawn Chen, L.~Guibas, Jitendra Malik, and S.~Savarese.
  2020.
\newblock \href {http://arxiv.org/abs/1905.07553} {Which tasks should be
  learned together in multi-task learning?}
\newblock In \emph{ICML}.

\bibitem[{Tang et~al.(2021)Tang, Pino, Wang, Ma, and Genzel}]{Tang2020AGM}
Yun Tang, Juan Pino, Changhan Wang, Xutai Ma, and Dmitriy Genzel. 2021.
\newblock \href {http://arxiv.org/abs/2010.11338} {A general multi-task
  learning framework to leverage text data for speech to text tasks}.
\newblock In \emph{ICASSP}.

\bibitem[{Vandenhende et~al.(2020)Vandenhende, Georgoulis, Gansbeke, Proesmans,
  Dai, and Gool}]{Vandenhende2020MultiTaskLF}
Simon Vandenhende, S.~Georgoulis, Wouter~Van Gansbeke, M.~Proesmans, Dengxin
  Dai, and L.~Gool. 2020.
\newblock \href {https://arxiv.org/pdf/2004.13379.pdf} {Multi-task learning for
  dense prediction tasks: A survey}.
\newblock \emph{arXiv: Computer Vision and Pattern Recognition}.

\bibitem[{Wang et~al.(2020{\natexlab{a}})Wang, Tang, Ma, Wu, Okhonko, and
  Pino}]{Wang2020fairseqSF}
C.~Wang, Y.~Tang, X.~Ma, A.~Wu, D.~Okhonko, and J.~Pino. 2020{\natexlab{a}}.
\newblock \href {http://arxiv.org/abs/2010.05171} {fairseq s2t: Fast
  speech-to-text modeling with fairseq}.
\newblock In \emph{AACL (demo)}.

\bibitem[{Wang et~al.(2020{\natexlab{b}})Wang, Wu, Liu, Yang, and
  Zhou}]{Wang2019BridgingTG}
Chengyi Wang, Yu~Wu, Shujie Liu, Zhenglu Yang, and Ming Zhou.
  2020{\natexlab{b}}.
\newblock \href {http://arxiv.org/abs/1909.07575} {Bridging the gap between
  pre-training and fine-tuning for end-to-end speech translation}.
\newblock In \emph{AAAI}.

\bibitem[{Weiss et~al.(2017)Weiss, Chorowski, Jaitly, Wu, and
  Chen}]{Weiss2017SequencetoSequenceMC}
Ron~J. Weiss, Jan Chorowski, Navdeep Jaitly, Yonghui Wu, and Zhifeng Chen.
  2017.
\newblock \href {https://arxiv.org/pdf/1703.08581.pdf} {Sequence-to-sequence
  models can directly translate foreign speech}.
\newblock In \emph{INTERSPEECH}.

\bibitem[{Zhang et~al.(2020)Zhang, Williams, Titov, and
  Sennrich}]{Zhang2020ImprovingMM}
Biao Zhang, Philip Williams, Ivan Titov, and Rico Sennrich. 2020.
\newblock \href {https://arxiv.org/pdf/2004.11867.pdf} {Improving massively
  multilingual neural machine translation and zero-shot translation}.
\newblock In \emph{ACL}.

\end{thebibliography}
\bibliographystyle{acl_natbib}

\newpage
\appendix
\section{Appendix}\label{sec:app}
The detailed experimental setup for ``MT'' and ``MT(Tuned)'' in \autoref{tab:mt} are described as below.

We trained MT models for each language pair in ``EN-DE'', ``EN-ES'', and ``EN-FR''. The training data is from WMT from different years, 2013 for ``EN-ES'', 2014 for ``EN-DE''  and 2016 for ``EN-FR''.  We use ``transformer\_wmt\_en\_de'' architecture from Fairseq. The models are with embedding size 512 and feed-forward layer dimension 2048. Both encoder and decoder are with 6 transformer layers. The input is phoneme sequence and output is SentencePiece sequence. The vocabularies are shared with the corresponding speech to text translation models. The models are optimized with Adam with learning rate equal to $0.001$. Beside experiments in \autoref{tab:mt}, the trained MT models are used to initialize the jointly trained models.

We further fine-tuned the ``MT'' models trained from WMT data to \textsc{MuST-C} data sets using source transcripts and target translation labels. No speech data is used. Similar to the ``MT'' models, Adam optimizer with learning rate equal to $0.001$ is used. The models are fine-tuned on the corresponding \textsc{MuST-C} data sets for 15 epochs and the checkpoints from the last 5 epochs are averaged for evaluation.  

\end{document}